\documentclass[conference]{IEEEtran}
\IEEEoverridecommandlockouts
\usepackage{cite}
\usepackage{amsmath,amssymb,amsfonts}
\usepackage{algorithm}
\usepackage{algorithmicx}
\usepackage{algpseudocode}
\usepackage{graphicx}
\usepackage{textcomp}
\usepackage{xcolor}
\usepackage{amsmath,epsfig}
\usepackage{multirow}
\usepackage{array}
\usepackage{graphicx}
\usepackage{subfigure}
\usepackage{booktabs}
\usepackage{bm}
\usepackage{array}

\def\BibTeX{{\rm B\kern-.05em{\sc i\kern-.025em b}\kern-.08em
    T\kern-.1667em\lower.7ex\hbox{E}\kern-.125emX}}

\begin{document}

\title{Adaptive Sparse and Monotonic Attention for Transformer-based Automatic Speech Recognition
	\thanks{${* }$ Corresponding Author: Jianzong Wang, jzwang@188.com}
}

\author{\IEEEauthorblockN{Chendong Zhao$^{\dagger \star \ddagger}$\thanks{$\ddagger$ Work done during an internship at Ping An Technology. }, Jianzong Wang$^{\dagger * }$, Wenqi Wei$^{\dagger}$, Xiaoyang Qu$^{\dagger}$, Haoqian Wang$^{\star}$, Jing Xiao$^{\dagger}$}
	\IEEEauthorblockA{$^{\dagger}$ \textit{Ping An Technology (Shenzhen) Co., Ltd., Shenzhen, China} \\
		{$^{\star}$ \textit{The Shenzhen International Graduate School, Tsinghua University, China}}
}}

\maketitle

\begin{abstract}
The Transformer architecture model, based on self-attention and multi-head attention, has achieved remarkable success in offline end-to-end Automatic Speech Recognition (ASR). However, self-attention and multi-head attention cannot be easily applied for streaming or online ASR. For self-attention in Transformer ASR, the softmax normalization function-based attention mechanism makes it impossible to highlight important speech information. For multi-head attention in Transformer ASR, it is not easy to model monotonic alignments in different heads. To overcome these two limits, we integrate sparse attention and monotonic attention into Transformer-based ASR. The sparse mechanism introduces a learned sparsity scheme to enable each self-attention structure to fit the corresponding head better. The monotonic attention deploys regularization to prune redundant heads for the multi-head attention structure. The experiments show that our method can effectively improve the attention mechanism on widely used benchmarks of speech recognition.
\end{abstract}


\section{Introduction}
\label{sec:intro}
Automatic speech recognition (ASR) has been well explored since it was first perceived. End-to-end ASR has seen large improvements in recent years. The great success of end-to-end paradigms for ASR should largely be attributed to several advanced models, such as Connectionist Temporal Classification (CTC) \cite{ASR_CTC},  Recurrent Neural Network Transducer (RNN-T) \cite{ARR_rnn_t}, and attention-based ASR \cite{ASR_Transformer,AttenASR_First,AttenASR_Listen_Spell}. 
The Transformer architecture has shown promising performance for end-to-end ASR~\cite{ic1,ic2,g2p,tsunoo2021streaming}, which combines an acoustic model, a language model, and an acoustic-to-text alignment mechanism. Unlike recurrent neural networks, Transformer can perform parallel training very well as it abandons the recurrent structure that has an inherently sequential nature~\cite{fedtts,rare,sv,hear}. Moreover, multi-head attention in the Transformer can significantly improve the network's performance because it can achieve dynamic embedding and capture the strong intrinsic structures~\cite{rform,mst,cai2021,cst}. However, although the Transformer shows expressive power for Natural Language Processing (NLP) tasks, there still remains room for exploiting it in speech tasks.

Transformer ASR methods~\cite{ic1,ic2} exploit the self-attention approach, where each self-attention layer in both encoder and decoder utilizes the multi-head structure to extract different input representations.
While it has been studied that some of the attention heads are useless in other tasks, we previously observed this phenomenon in Transformer ASR as well, where the attention function is similar to an identity mapping.
This issue motivates us to find out and address the redundant calculations in such self-attention.
So in this work, we propose to remove the redundant informations without degrading the overall performance.
For Transformer-based ASR model, one fundamental problem is that the softmax normalization function-based attention mechanism makes it impossible to highlight important speech information. In the attention mechanism, softmax is often used to calculate the attention weight, and its output is always dense, and non-negative~\cite{c1,SparseAtt_SparseMax, SparseAtt_Regularized}.
Dense attention means that all inputs always contribute a little bit to the decision, which leads to a waste of attention. Besides, the attention score distribution becomes flat as the input length increases. Therefore, it is difficult to focus on the speech's critical information, especially for a long speech. Considering that dense attention cannot highlight important information in a long speech, we try to use sparse transformation to achieve sparse attention distribution. Inspired by \cite{SparseAtt_Adaptive}, we first apply the adaptive $\alpha$-entmax to the ASR task. Specifically, we replace the softmax in each self-attention with $\alpha$-entmax, and maintain the learnability of $\alpha$.

For attention-based ASR model, another fundamental challenge is the accurate alignment between the input speeches and output texts~\cite{c2}. Monotonicity can keep the marginalization of the alignment between text output and speech input tractable. But, the monotonic multi-head alignment can not make all heads contribute to the final predictions. We proposed an adaptive monotonic multi-head attention with a regularization constraint.
In our experiments, we found some techniques which highlight attention scores of important parts of the speech, offering better alignment accuracy. In practice, searching for the best function is time-consuming. To tackle this problem, we introduce an approach based on automatically adaptive attention. Contributive to such alignment, the trained model is more focusing and accurate of multiple attentions.
In this way, it can adaptively eliminate the redundant heads and let every head learn alignments properly.

Overall, our contributions are as follows. Firstly, we unify the sparse attention and monotonic attention into ASR tasks. Secondly, we propose an adaptive method to obtain a unique $\alpha$ parameter for each self-attention in Transformer. Third, we deployed a regularized monotonic multi-head alignment scheme for acoustic-to-text alignment.

\section{Related Works and Preliminary}
\subsection{Sparse Attention}
Recent works~\cite{SparseAtt_Structured,SparseAtt_SparseMAP,SparseAtt_Constrained_Attention}
based on the attention function propose to use sparse normalizing such as sparsemax \cite{SparseAtt_SparseMax}. The sparsemax can omit items with small probability and thus can distribute the probability mass on outputs with higher probabilities. It improves the performance and interpretability to a certain extent \cite{SparseAtt_Regularized}. However, each head of the multi-head attention mechanism used in Transformer is designed for capturing characteristics from a different point of view. For example, while the heads at the first layer of encoder focus on feature extraction, the heads in the last layer of decoder focus on classification of results. Intuitively the attention function of the two scenarios should be different. So the method of replacing all attention mechanisms with sparsemax may not suit for Transformer \cite{SparseAtt_SyntacticTrees}. In the field of NLP, there are many breakthroughs in Transformer regarding sparsity. From the perspective of Tsallis entropy \cite{SparseAtt_polytropes}, \cite{SparseAtt_Seq2Seq} unify all transformation functions into the form of $\alpha$-entmax. 
When $\alpha$ = 1, $\alpha$-entmax is a softmax function. For any $\alpha$ $>$ 1, $\alpha$-entmax is the sparsemax function. Moreover, in order to make the $\alpha$-entmax fit different heads for different tasks, \cite{SparseAtt_Adaptive} introduces an adaptive method to enable the parameter $\alpha$ to be updated. 

\subsection{Monotonic Attention}
Monotonic attention technique~\cite{mono} aims to restrict the transition of attention in a past-to-now pipeline. Generally speaking, monotonic attention is adopted to reduce the complexity of decoding.
In strong context-related tasks as natural language understanding, a fully attention function is commonly applied to both input and output sequences, because the word-order or the part-of-speech is less important.  
But it may not work well for ASR tasks because the speech waveforms and target text sequence are required to be temporally monotonic.
For enhancing the alignment ability between the waveform and text sequence, \cite{kim2017joint} incorporate the monotonic attention together with the connectionist temporal classification objective in a multi-task learning diagram by using the same encoder together.
This recipe becomes the standard optimizing pipeline for most attention-based models~\cite{1,2,3}. In~\cite{AttenASR_MoChA}, a monotonic chunkwise attention (MoChA) approach was introduced for a streaming attention. It first splitted the encoder output latents into rather short-length chunks and then applied the soft attention function on these small chunks. Furthermore, an adaptive-length chunk approach was proposed in~\cite{45}. Specifically, monotonic infinite lookback (MILK) attention~\cite{46} was applied to consider the overall acoustic features preceding to learn an adaptive latency-quality trade-off schedule jointly.

\subsection{Attention-based ASR}
Although attention-based architectures\cite{att, AttenASR_First,AttenASR_Listen_Spell} have achieved remarkable success in offline end-to-end ASR systems, they cannot easily be applied for the online or streaming ASR systems. Several optimization paradigms are proposed to overcome this limit. Monotonic multi-head attention\cite{AttenASR_MMA} extends monotonicity to the multi-head to extract useful representations and complex alignment. For Monotonic multi-head attention \cite{AttenASR_MMA}, HeadDrop\cite{AttenASR_Kyoto_HeadDrop} is proposed to eliminate the redundant head. MoChA \cite{AttenASR_MoChA} bridges the gap between monotonic attention and soft attention.
As the online speech recognizing becoming more and more important, it also brings a drawback of additional designings in the network and training criterias, and precision degradations.
As the Transformer's size is large, a compressed structure \cite{AttenASR_Kyoto_Shared} is proposed. Conformer \cite{AttenASR_ASR_Conformer} is a convolution-augmented Transformer for speech recognition. In triggered attention \cite{AttenASR_TriggeredAtt}, a CTC task is trained to learn the alignment that triggers the attention decoding. The work in \cite{AttenASR_SA_Aligner} use chunk-hopping for a latency-control end-to-end ASR.

\subsection{Transformer ASR Model Architecture}

As shown in Figure 1, our network architecture is based on a Transformer-based ASR model, based on an encoder and a decoder. Let us denote the input sequence as $x=\{x_1,...,x_T\}$ with T being the length of the frame sequence. The corresponding encoder states are denoted as $h=\{h_1,...,h_S\}$. Let the output sequence as $y=\{y_1,...,y_U\}$ with $U$ being the length of the character sequence. 
The encoder first processes the input with a down-sampling convolution layer and stacks with several encoder blocks. The encoder processes sequential speech features, which are 80-dimensional log-mel spectral energies. The down-sampling convolution layer uses a stride of 2, kernels of $3\times3$, followed by the ReLU activation function. The down-sampling convolution is able to extract useful encodings, shorten acoustic representations' length, and promote faster alignments in the decoding. 
Main modules of Transformer contain position encoding, multi-head self-attention calculation, linear, and softmax, point-wise feed-forward network, residual connections, layer normalization. Here, we emphasize the concept of self-attention and multi-head attention. Self-attention is designed to learn the internal dependencies by computing the representation of a single sequence.
The Transformer employs the scaled dot-product self-attention formulated as:
\begin{align}
	head_i=softmax(\frac{Q_{i}K_{i}^T}{\sqrt{d/h}})V_i\: \: (i=1,2,...,h),\\ 
	where \:\: Q_i=XW_i^q,K=X_1W_i^k,V_i=XW_i^v.
\end{align}
Here, $h$ means the number of heads of multi-head attention. As the dimension of $X$ in Transformer has the same dimension as $d_{model}$, the project matrices are $W_i^q \in \mathbb{R}^{{d_{model}} \times d_q}$, $W_i^k \in \mathbb{R}^{d_{model} \times d_k}$, $W_i^v \in \mathbb{R}^{d_{model}\times d_v}$. Note that $d_q=d_k=d_v=d_{model}/h$. 
The output of self-attention $O$ is concatenated by multi-head mechanism, which is formulated as follows
\begin{equation}
	O= concat(head_1,head_2,...,head_h)W^{O}.
\end{equation}
$W^O \in \mathbb{R}^{hd_{v} \times d_{model}}$ and $concat(\cdot)$ means concatenation. Besides, it deploys the sinusoidal positional encoding scheme, which achieves the attention calculation to adapt well on different speech length.

\section{Approach}
\label{sec:format}
In order to systematically prove the performance of adaptive sparse transformer, we designed multiple transformer structures based on different sparse functions, including sparse transformer, 1.5-entmax transformer, and adaptive $\alpha$-entmax transformer.
Our innovations are as follows: firstly, we use different sparse functions to replace the softmax function in self-attention, and use adaptive sparse methods to each self-attention gets its own degree of sparsity. Secondly, we propose a regularized monotonic multi-head attention.
Our innovations lie in two components: an adaptive sparse multi-head self-attention illustrated in Sec.~\ref{sec:AdaptiveSparse} and an adaptive monotonic multi-head attention illustrated in Sec.~\ref{sec:AdaptiveMonotonic}. Figure 1 shows how to integrate these two core components into the baseline.

\begin{figure}[t]  
	\centering
	\includegraphics[scale=0.5]{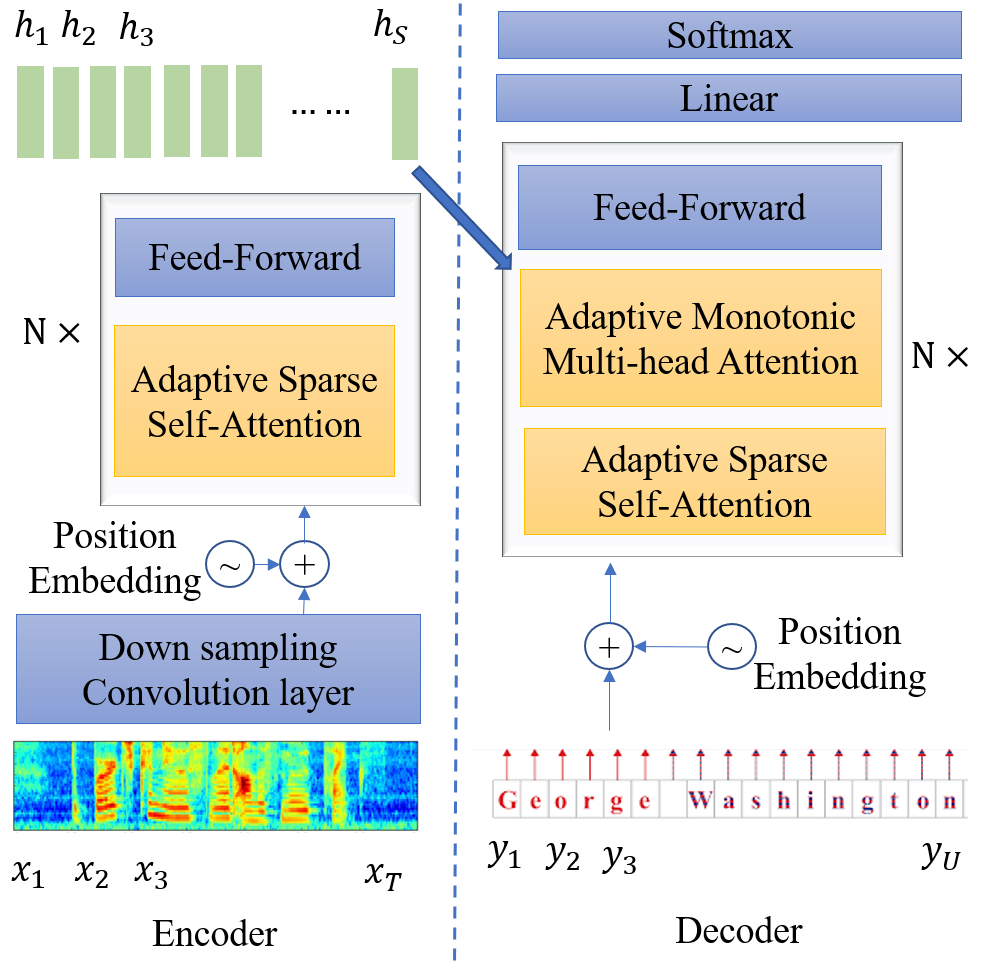}
	\caption{The diagram of our model architecture. Note that the residual connections and layer normalization are omitted.}
	\label{1}
\end{figure}

\subsection{Adaptive Sparse Multi-head Self-Attention}\label{sec:AdaptiveSparse}
As shown in Figure 2, we introduce several sparse functions in self-attention to prevent the disadvantage of softmax.
The first idea is to substitute softmax with sparsemax for every independent head of the Transformer, which can totally omit the attention of low probability and distribute probability mass only on high-interest elements. It is achieved by computing the Euclidean projection of the input vector $\textbf{z}$ onto the probability simplex p:
\begin{equation}
	Sparsemax(\textbf{z})= arg\min\limits_{\textbf{p}\in \Delta^{J}}\frac{1}{2}||\textbf{z}-\textbf{p}||^{2}_{2},
	\label{sparsemax}
\end{equation}
where $\Delta^{J}={\textbf{p}\in R^{J}|\textbf{p}\leq0,\sum\limits_{j}^{J}p_{j}=1}$ is J-dimension simplex.
To overcome the potential issue of softmax, we propose to treat $\alpha$ values of each attention head differently, where some heads may train to be informative sparser, and others may become similar its original form.
We also propose to treat the $\alpha$ values as learnable variables, optimized via back-propagation calculations along with other network parameters.
In~\cite{SparseAtt_SparseMax}, the sparsity of sparsemax is proved and a closed-form solution is given:
\begin{equation}
	\textbf{p}^{*} = sparsemax(\textbf{z}) = [\textbf{z}-\tau]_{+},
\end{equation}
where $[\cdot]_{+}$ denotes function $max(\cdot,0)$, and $\tau$ is the Lagrange multiplier in the context of the constraint $\sum_j(p_j)=1$. Here $\tau$ can also be interpreted as a threshold under which the corresponding probability will be set to zero. Next, we extend the concept of sparse attention in the multi-head architecture. The core change of this architecture is we use $\alpha\text{-entmax}$ \cite{SparseAtt_Seq2Seq} to introduce sparsity into attention mechanism of sparse self-attention. For the three mentioned transformations (softmax, sparsemax, $\alpha$-entmax), the relationship among them can be analyzed from the perspective of entropy:
\begin{equation}
	\alpha\text{-entmax}(\textbf{z}) = arg \max\limits_{\textbf{p}\in \Delta^{J}}\textbf{p}^{T}\textbf{z}+H_{\alpha}^{T}(\textbf{p}).
	\label{a-entmax}
\end{equation}
$H_{\alpha}^{T}(\textbf{p})$ is the Tsallis $\alpha$-entropy family definition with the following definition:
\begin{equation}
	H_{\alpha}^{T}(\textbf{p}) = 
	\begin{cases}
		
		\frac{1}{\alpha(\alpha-1)}\sum\limits_{i}(p_j-p_{j}^{\alpha}) & \alpha \neq 1\\
		
		-\sum_jp_jlogp_j & \alpha=1
		
	\end{cases}
	\label{a-entropy}
\end{equation}
$\alpha$ here is a hyperparameter. For the cases $\alpha =1$ and $\alpha =2$, the corresponding $\alpha$-entropies are Shannon and Gini entropy while the corresponding $\alpha$-entmaxes are equal to softmax and sparsemax respectively. Thus it establishes the connection between softmax and sparsemax while also throwing lights on a medium model between these two transformations.
\cite{SparseAtt_Seq2Seq} proves the uniqueness of $\tau$ and proposed an accurate method for calculating $\alpha$-entmax when $\alpha$ = 1.5 (also called 1.5-entmax). \cite{ASR_Fenchel_Young} proposed an extended iterative bisection approach that is applicable for $\alpha$-entmax.
It is rational to set the $\alpha$ to different values in Transformer for better adaptation to different heads in different layers. 

Inspired by \cite{SparseAtt_Adaptive}, we apply an adaptive $\alpha$-entmax attention method in the multi-head attention mechanism for ASR tasks. $\alpha$-entmax is a convex optimization problem. When $\alpha$ is set as a variable but not a hyperparameter, we have $p^{*}=\alpha-entmax(\textbf{z},\alpha)$. It is not trivial to derivate $\frac{\partial \alpha-entmax(\textbf{z},\alpha)}{\partial \alpha}$ from the Lagrangian and optimality conditions by taking the derivative with respect to \textbf{z} in \cite{SparseAtt_SparseMax}. Due to current deep neural network frameworks can not take the derivatives for this optimization problem automatically. Thus, we tackle this problem from the perspective of the solution. It is trivial that for $j \notin S$, $p^{*}_{j}=0$ which is a constant thus $\frac{\partial p^{*}_{j}}{\partial \alpha}=0$. To simplify the gradient for $i \in S$, we use $\overline{p}^{*}$ and $\overline{\textbf{z}}$ to denote the corresponding vectors whose indices are in the support $S$.
Finally we can solve the component of the Jacobian.

\begin{figure}[t]  
	\centering
	\includegraphics[scale=0.4]{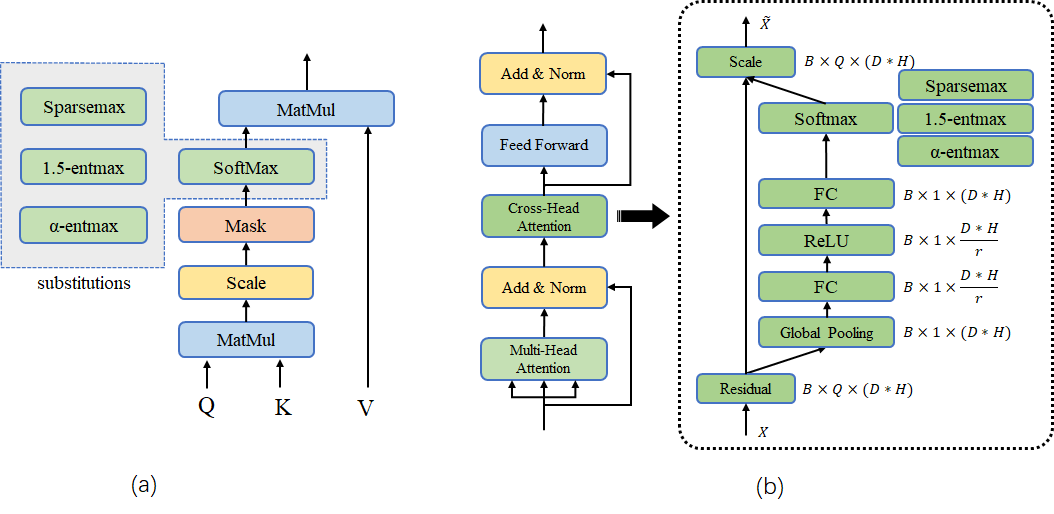}
	\caption{Illustrations of the proposed adaptive sparse attention.}
	\label{1}
\end{figure}
\subsection{Adaptive Monotonic Multi-head Attention}\label{sec:AdaptiveMonotonic}
For end-to-end ASR, another fundamental challenge is the alignment between the input sequence and output sequence. CTC \cite{ASR_CTC} and  RNN-Transducer \cite{ARR_rnn_t} are monotonic, so that the alignment track is able to be marginal. But the attention approach results in non-monotonic alignment, which may diminish the quality of speech recognition. The hard monotonic attention mechanism was introduced used for streaming time-sync decoding with attention architecture ASR.
For the hard monotonic attention mechanism, the process is shown as follows. For the decoding time-step i, the attention method begins inspecting entries starting at the preceding output time-step, referred to as $t_{i-1}$. It then computes an energy scalar $e_{i,j}$ based on a monotonic energy function $MonotonicEnergy(\cdot)$. This function refers the $(i-1)_{th}$ output state and encoding features $h_j$ as inputs, where $j=t_{i-1},t_{i-1}+1,...$. The energy scalar is computed as follows:

\begin{table*}[t]
	\centering
	\caption{Results of different Max functions (CER (\%) in AISHELL, WER (\%) in WSJ), T represents softmax temperature. Numbers in the last three rows represent sparsing only encoder / only decoder / both.}
	\label{Tab03}
	\resizebox{0.9\linewidth}{18mm}{
		\begin{tabular}{ccccccc}
			\toprule[1pt]
			\multirow{2}{*}{Method} & \multicolumn{3}{c}{Sparse Attention} & \multicolumn{3}{c}{Sparse Attention and Monotonic Attention} \\ \cline{2-7}
			& ~~~AISHELL-Dev~~~ & ~~~AISHELL-Test~~~  &  ~~~WSJ-eval93~~~ &  ~~~AISHELL-Dev~~~  & ~~~AISHELL-Test~~~ & ~~~WSJ-eval93~~~ \\ \midrule[1pt]
			softmax & 8.60 & 9.33 & 9.58 & 8.23 & 9.00 & 9.36      \\
			softmax(T=0.75) & 8.54 & 9.03 & 9.34 & 8.43 & 8.91 & 9.07 \\
			softmax(T=0.5) & 7.80 & 8.64 & 8.80 & 7.73 & 8.59 & 8.64 \\
			softmax(T=0.25) & 8.07 & 8.91 & 9.05 & 7.87 & 8.69 & 8.91 \\
			\midrule[1pt]
			sparsemax  & 8.91 / 8.99 / 8.98 & 9.56 / 9.53 / 9.55 & 9.69 / 9.64 / 9.62 & 8.64 / 8.58 / 8.64 & 9.13 / 9.29 / 9.06 & 9.45 / 9.38 / 9.34 \\
			1.5-entmax &7.69 / 7.71 / 7.69 & 8.57 / 8.62 / \textbf{8.49} & 8.47 / 8.58 / 8.56 & 7.68 / 7.67 / 7.68 & 8.51 / 8.59 / 8.47 & 8.49 / 8.55 / 8.39 \\
			$\alpha$-entmax (adaptive) & 7.67 / 7.71 / \textbf{7.65} & 8.54 / 8.60 / 8.53 & \textbf{8.35} / 8.47 / 8.38 & 7.71 / 7.69 / \textbf{7.64} & 8.51 / 8.52 / \textbf{8.45} & 8.38 / 8.49 / \textbf{8.37} \\
			\bottomrule[1pt]
	\end{tabular}}
\end{table*}

\begin{equation}
	e_{i,j} = MonotonicEnergy(s_{i-1},h_j).
\end{equation}
The detail of monotonic energy function can refer to \cite{AttenASR_MMA}. Then the energy values $e_{i,j}$ are passed into a sigmoid function $\sigma(\cdot)$ to produce  selection probability $p_{i,j}$ formulated as
\begin{equation}
	p_{i,j} =\sigma(e_{i,j}).
\end{equation}
Then, the selection probability $p_{i,j}$ is used to produce a discrete decision variable $z_{i,j}$, which is formulated as
\begin{align}
	z_{i,j} \sim Bernoulli(p_{i,j}).
\end{align}
Whenever $p_{i,}$ is satisfied, the $z_{i,j}$ is stimulated (i.e., value approaching 1). Once $z_{i,j} =1$, the model stops and set $t_i=j$ and $c_i = h_{t_i}$. 

In contrast to a single attention head in the recurrent sequence-to-sequence model, multi-head mechanism is capable of learning contexts with diverse paces between input and output sequence. Besides, the multi-head can learn complex alignment. In specific, the results of each head can be regarded as corresponding to some specific frames in the speech, and different heads are related to each other. While some heads can detect boundaries, other heads can capture strong intrinsic speech structures.

Dense attention mechanisms can fail to capture or utilize the strong intrinsic structures present in speech. By taking advantage of the dependencies between specific frames that different heads focus on, we can emphasize the key information in the speech and highlight the characteristic representation of these speech frames. 
However, there is no gurantee that multiple heads are all contributive to the overall context learning. Besides, the monotonic multi-head alignment can prevent delayed state decoding caused by multi-heads for boundary detection. Every monotonic attention head must learn alignments properly.
It is necessary to prune some redundant monotonic attention heads. Thus, to enhance accordance among multi-heads on token boundary discovery, we use L1 regularization to eliminate information-redundant heads in shallow layers.

\section{Experiment}
\label{sec:format}

In this section, we compare our method with the standard Transformer architecture of the softmax transform. Also, we compare with other two method variants for a fair comparasion, which explore different normalizing transformations:

\noindent\textbf{1.5-entmax}: Set fixed value $\alpha$ = 1.5 for all multi-heads of a sparse ent-max Transformer. This approach is considered to be novel, since 1.5-entmax had solely been introduced for recurrent ASR methods, but not explored in Transformer before. The major difference is, assigning sparse ent-max is only one module of seq2seq method, but being an integral component of all the Transformer modules.

\noindent\textbf{$\alpha$-entmax}: To be benefit from an adaptive value of the sparse ent-max transformation, a learnable $\alpha_{i,j}^t$ is assigned for every attention head.

\subsection{Dataset and Setup}
\noindent\textbf{Datasets}: We systematically evaluated model performance on AISHELL-1~\cite{ASR_Aishell}
and WSJ. AISHELL-1 is a Chinese corpus, which consists of 400 speakers and almost 170 hours of recorded utterances. The corpus utterances are splitted to training, develop and test datasets, where the training set consists of 120,098 speeches of 340 speakers; the develop set consists of 14,326 speeches from 40 speakers and test set consists of 7,176 utterances from 20 speakers.
In the experiments of AISHELL-1, we do not use language models.
For the WSJ dataset, we utilize eval-92 for evaluation and eval-93 for test. We use transcribed text to train a N-gram language model.

\noindent\textbf{Metrics}: Word error rate (WER) and character error rate (CER) were applied as the evaluation metrics in the experiments.
WER is calculated by the portion of words whose recognized text-formulations were not same with the ground-truth labels.
CER is calculated based on the Levenshtein distance between the recognized and the ground-truth characters, then divided by the absolute sequence length.
Both metrics follows that the lower of values, the better the performance.

\noindent\textbf{Training Details}:
As mentioned before, the input speech waveform is firstly processed to deep features of 80-dimensional filterbanks, where the hop size is 10ms and the window size is 25ms. Also, we conduct on temporal and speaker differences subtraction and bias normalization. The architecture setting includes the encoder block number $N_e  = 12$, the decoder block number $N_d = 6$ and the feed-forward network of $d_{ff} = 1024 $. We also keep feature dimension $d_{model} = 512 $ and attention heads $h = 8$ same as the baseline. We used the Adam optimizer with $\beta_1$ = 0.9, $\beta_2$ =0.98 and the warmup steps is 25000. The dropout is 0.1. For decoding, we use  beam search with a beam size of 5. We only changed the softmax function in the self-attention in the encoder and add adaptive monotonic multi-head attention in the decoder.
For comparison methods, we adopt a RNN-Transducer.
The model is a 4-layer bidirectional long short-term memory model of 320 hidden units, each direction account for encoding, and a 2-layer uni-directional long short-term memory model of 512 hidden units as recognization module. This architecture considers both the acoustic and linguistic features, and projects the prediction of non-linear activation function to softmax.

\subsection{Overall Results}

\begin{table}[t] 
	\centering
	\caption{WER (\%) comparisons on WSJ eval93}
	\label{t2}
	\resizebox{0.8\linewidth}{14mm}{
		\begin{tabular}{ccc}  
			\toprule[1pt]
			~~~~~~~Method~~~~~~~  & ~~~~~~WER~~~~~~ \\
			\hline
			Transformer & 9.39  \\
			\hline
			wav2letter \cite{ASR_wav2letter} & 9.5  \\
			Jasper \cite{ASR_Jasper} & 9.9 \\
			\hline
			Explicit Sparse Transformer \cite{SparseAtt_Explicit} & 12.98  \\
			Factorized Attention Transformer \cite{SparseAtt_FactorizedAttention} & 8.97  \\
			Combiner \cite{SparseAtt_Combiner}& 9.32  \\
			\hline
			Proposed method & 8.22 \\
			\bottomrule[1pt]
	\end{tabular}}
\end{table}
\begin{table}[t] 
	\centering  
	\caption{CER (\%) comparisons on on AISHELL-1}
	\label{t3}
	\resizebox{0.8\linewidth}{13mm}{
		\begin{tabular}{cccc}  
			\toprule[1pt]
			~~~~~~Method~~~~~~  & ~~~~~~Dev~~~~~~ & ~~~~~~Test~~~~~~ \\
			\hline
			Transformer & 8.47 & 9.32 \\
			\hline
			RNN-T & 10.13 & 11.82 \\
			LAS \cite{AttenASR_Listen_Spell} & - & 10.56 \\
			SA-T \cite{ARR_rnn_t}  & 8.30 & 9.30 \\
			Sync-Transformer \cite{ASR_sytransformer} & 7.91 & 8.91 \\
			\hline
			Proposed method & 7.58 & 8.40 \\
			\bottomrule[1pt]
	\end{tabular}}
\end{table}

The results are shown in Table 1. When only sparse transforms are added to the Transformer encoder and decoder, both have better performance. Among them, adding to both the encoder and decoder have achieved better results. We speculate that this is because the sparse transform enhances the discrimination of the acoustic features, and the discrimination of the text information is less important than the comparison. In addition, the model that completely replaces the sparse transform achieves the best results.
The method of sparsity is significantly improved compared to the method of changing the softmax temperature. Low softmax temperatures can make the distribution of softmax results steeper, thereby highlighting the expression of some information, but the non-negative output of softmax is still not changed, which still leads to waste of attention. It is worth noting that a too low temperature will cause the model to focus more on a certain part of the information and ignore other important information, leading to a decline in the results.

\begin{table}[t] 
	\centering  
	\caption{Sparsity comparisons (WER \%) on LibriSpeech Corpus.}
	\label{t4}
	\resizebox{0.8\linewidth}{9mm}{
		\begin{tabular}{ccccc}  
			\toprule[1pt]
			~~~Method~~~ &~~~dev-clean~~~ & ~~~dev~~~ & ~~~test-clean~~~ &~~~test~~~ \\ 
			\hline
			softmax & 3.64 & 11.89 & 3.86 & 11.95\\ 
			\hline
			sparsemax & 3.88 & 12.01 & 4.00 & 12.19 \\
			1.5-entmax & 3.48 & 10.43 & 3.79 & 11.76 \\
			$\alpha$-entmax & 3.36 & 10.29 & 3.70 & 11.64 \\
			\bottomrule[1pt]    
	\end{tabular}}
\end{table}

In self-attention, replacing the softmax function with adaptive $\alpha$-entmax all reduces WER, which shows that adding sparseness is beneficial to the performance of the model. However, sparsemax caused a decrease in performance, which shows that excessive sparseness can cause loss of  information. This demonstrates that excessive sparseness cannot highlight the key information in the speech, and only a properly designed adaptive sparse can effectively highlight the key information. We combine the adaptive $\alpha$-entmax in self-attention and report the best result in Table~\ref{t2},~\ref{t3}.

\subsection{The Analysis of Adaptively Sparse Attention}

\begin{figure}[t]
	\centering
	\includegraphics[scale=0.4]{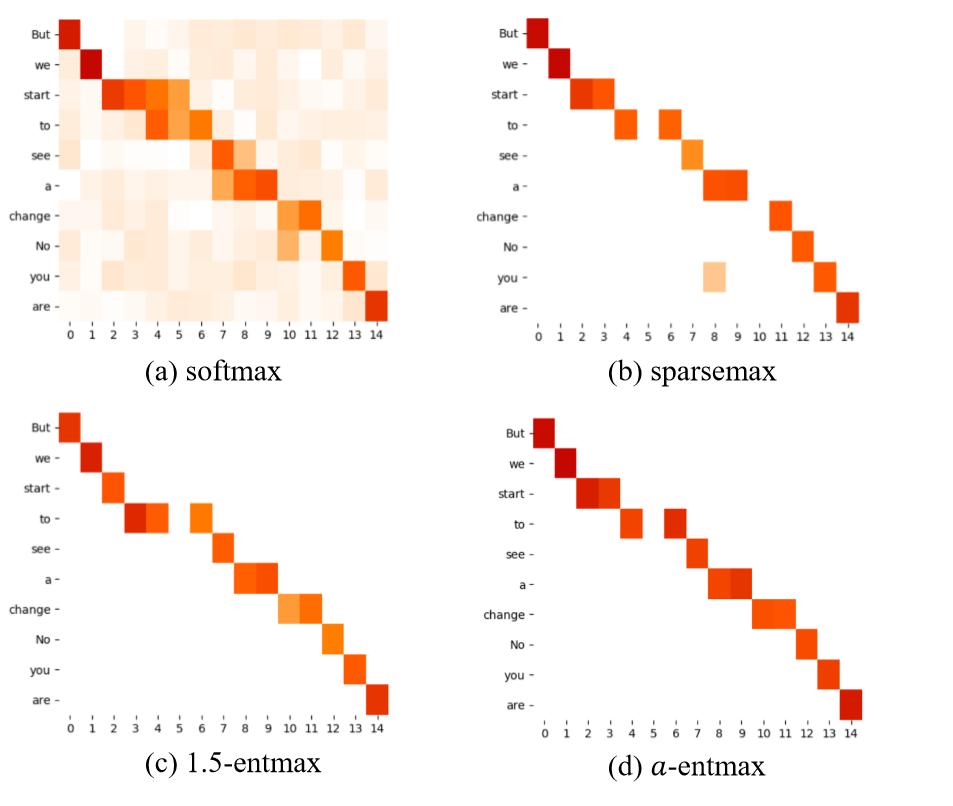}
	\caption{Attention map of different transformations in sparse self-attention.}
	\label{att_map}
\end{figure}

In this section, we retain the monotonic attention and study the effect of different sparse transformations in the sparse self-attention.
In Table 1, we keep the self-attention module in its most original state (softmax) and introduce it into the cross-head attention block for experiments. The third to sixth rows in Table 1 represent different max functions in the cross-head attention block. Due to the small number of attention heads, too sparse max function (sparsemax) will lead to the loss of information. The max function with insufficient sparsity does not play a sparsity role. It is worth mentioning that the adaptive $\alpha$-entmax function tends to $\alpha$=1. To further investigate the proposed method on more challenging data sets, we also conduct ablation studies on English LibriSpeech dataset in Table~\ref{t4}.
In Figure \ref{att_map}, we pose an example with different sparse transformations in sparse self-attention. Compared with softmax, the sparsemax function can produce competitive sparse output, but it will miss some output information. 1.5-entmax can produce more complete output results, thus achieving better performance. $\alpha$-entmax enables each self-attention module to use a specific $\alpha$ according to different emphases, thus obtaining the best performance. We can see that $\alpha$-entmax does not only avoid the waste of attention scores, but also effectively highlights some important frames in speech. The deepening of the color means that it has a higher attention score.

\begin{figure}[t]
	\centering
	\includegraphics[scale=0.45]{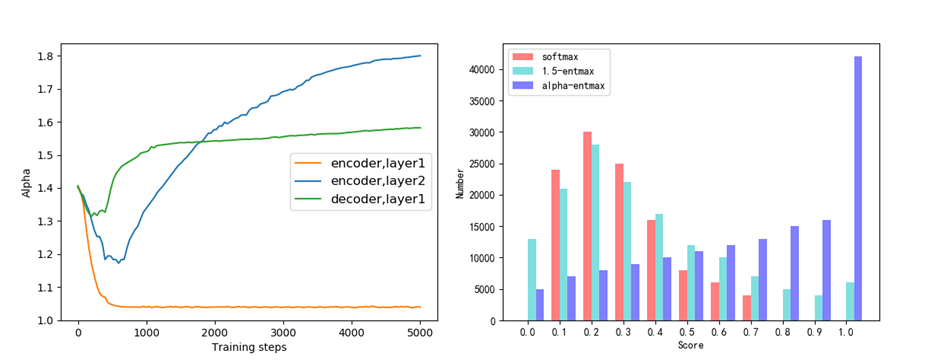}
	\caption{(a) The change of $\alpha$ in sparse self-attention during training, (b) The statistics of the final score of the various attention mechanisms.}
	\label{parameter_change}
\end{figure}

\begin{table}[t]
	\centering
	\caption{Ablation studies about the L1 regularization on monotonic attention.}
	\resizebox{0.9\linewidth}{11mm}{
		\begin{tabular}{c|cc|cc}
			\toprule[1pt]
			\multirow{2}{*}{~~~~~Method~~~~~} & \multicolumn{2}{c}{~~~~~AISHELL-Test~~~~~} & \multicolumn{2}{c}{~~~~~WSJ-eval93~~~~~} \\ \cline{2-5}
			& \textbf{reg.} & \textbf{w/o reg.}  &  \textbf{reg.} &  \textbf{w/o reg.} \\ \midrule
			softmax & 9.00 & 9.29 & 9.36 & 9.43   \\
			sparsemax & 9.13 & 9.32 & 9.45 & 9.47  \\  
			$\alpha$-entmax & 8.49 & 8.50 & 8.37 & 8.35 \\
			\bottomrule[1pt]
	\end{tabular}}
\end{table}

In Figure \ref{parameter_change}(a), we tracked the change of the $\alpha$ parameter. It can be seen that at the beginning of the training, the $\alpha$ parameter has decreased to a certain degree, which shows that it is desirable to use a dense output similar to softmax in the early stage of training. Especially for the encoder part, the closer to the feature input, the closer the value of $\alpha$ is to 1, which shows that the softmax function can effectively enhance the distinguishability of features. 
After a certain amount of training, except for the first layer of the encoder, the other parts move closer to the sparse direction, which shows that the method of initializing $\alpha$ cannot determine the degree of sparseness in advance, and to some extent, it shows the method of fixing $\alpha$ as a learnable parameter is suitable for multi-head attention. In Figure 3(b), we visualized the final output attention score of the cross-head attention, and we can see that softmax generates a lot of low score attention due to lack of sparsity. The structure we proposed can produce more high scores to avoid the waste of attention scores.

\subsection{The Analysis of Adaptively monotonic Attention}
In Table 1, comparing with the sparse attention alone, the preformance of sparsemax methods is improved by averaging -0.17\% with the help of monotonic attention. Furthermore, the monotonic attention is also complementary to the sparsemax and entmax methods. Test results are enhanced by -0.13\% CER on AISHELL and -0.07\% WER on WSJ, respectively. Results show that the monotonic attention can cooperate with the sparse attention to improve the performance further.

We further experiment the L1 regularization on the monotonic attention. As shown in Table 5, with the help of the regularization, all max function methods gain improvements both on AISHELL-1 and WSJ. The gap between w/o \textbf{reg.} and the proposed can be reduced from 9.29\% to 9.00\% on AISHELL-1 and 9.43\% to 9.36\% on WSJ for original softmax version. We applied adaptive sparse where $\alpha$ leading to 1.0, the monotonic multi-head alignment can prevent delayed token generation caused by such heads for boundary detection.

\vspace{3mm}
\section{Conclusion}

This paper integrates an adaptive sparse attention mechanism and an adaptive monotonic attention mechanism into a Transformer-based end-to-end ASR model. The adaptive sparse attention mechanism is implemented in multi-head self-attention to highlight important informations in the input speech by replacing the softmax function with $\alpha$-entmax. Besides, the adaptive monotonic multi-head attention is implemented by adding a regularized term. The experimental results prove that our method can focus more attention on the more important information and achieve the best results.
The adaptive mechanism bases only on gradient-based optimization, searching for per-head sparsity and computing $\alpha$-entmax.

\vspace{3mm}
\section{Acknowledgment}
This paper is supported by the Key Research and Development Program of Guangdong Province under grant No.2021B0101400003. Corresponding author is Jianzong Wang from Ping An Technology (Shenzhen) Co., Ltd (jzwang@188.com).

\vfill
\pagebreak

\bibliographystyle{IEEEtran}
\bibliography{bib/mybib,bib/AttentionASR,bib/SparseAttention}

\end{document}